\documentclass[11pt,theapa]{article}
\usepackage{theapa, rawfonts} 

\usepackage{xcolor}
\usepackage[T1]{fontenc}
\usepackage[latin9]{inputenc}
\usepackage{graphicx}
\usepackage{placeins}
\usepackage{float}
\usepackage{enumerate}
 \usepackage{url}

\usepackage{amsmath}
\usepackage{booktabs}

\usepackage[amsthm]{ntheorem}
\usepackage{todonotes}
\theoremstyle{plain}
\theoremheaderfont{\bfseries}
\theorembodyfont{\normalfont}
\theoremseparator{.} 
\newtheorem{question}{Question}
\newtheorem{case}{Example}

\makeatletter
\let\cl@chapter\undefined
\makeatletter
\usepackage{cleveref}

\definecolor{asparagus}{rgb}{0.53, 0.66, 0.42}
\definecolor{blue-violet}{rgb}{0.54, 0.17, 0.89}
\usepackage{xcolor}

\newcommand{\vulnerability}{{low impulse control}} 

\begin{document}

\title{Against Algorithmic Exploitation of Human Vulnerabilities}
\author{Inga~Str{\"u}mke  \and
       Marija Slavkovik\and Clemens Stachl 
     }




\maketitle

\begin{abstract}
Decisions such as which movie to watch next, which song to listen to, or which product to buy online, 
are increasingly influenced by recommender systems and user models that incorporate information on users' past behaviours, preferences, and digitally created content. Machine learning models that enable recommendations and that are trained on user data may unintentionally leverage information on human characteristics that are considered vulnerabilities, such as depression, young age, or gambling addiction. The use of algorithmic decisions based on latent vulnerable state representations could be considered manipulative and could have a deteriorating impact on the condition of vulnerable individuals. In this paper, we are concerned with the problem of machine learning models inadvertently modelling vulnerabilities, and want to raise awareness for this issue to be considered in legislation and AI ethics. Hence, we define and describe common vulnerabilities, and illustrate cases where they are likely to play a role in algorithmic decision-making. We propose a set of requirements for methods to detect the potential for vulnerability modelling, detect whether vulnerable groups are treated differently by a model, and detect whether a model has created an internal representation of vulnerability. We conclude that explainable artificial intelligence methods may be necessary for detecting vulnerability exploitation by machine learning-based recommendation systems.
\end{abstract}
\section{Introduction}
 Millions of people rely on recommender systems daily. Examples of recommendation based activities include listening to music via streaming services, shopping for products and services online, and browsing through social media streams. What we can choose from in the context of these activities is determined by algorithmic decision architectures that are optimised for a specified target such as maximising engagement, maximising the click through rate\footnote{\url{https://en.wikipedia.org/wiki/Click-through_rate}}, or increasing the sales. To enable these functionalities, recommender systems employ increasingly sophisticated user behaviour models that incorporate information on users' past behaviours, preferences, interactions and created content \cite{SlavkovikSPA21}. 

Machine learning (ML) models that enable recommendations and that are trained on user data may unintentionally leverage information on human characteristics that are considered vulnerabilities (e.g., depression, young age, gambling addiction). In simple terms, an ML model may include a parameterised representation of a person's vulnerability. Algorithmic decisions that are based on models using such representations could be considered manipulative, because the use of algorithmic decisions based on representations of vulnerabilities might have a deteriorating impact on the vulnerability condition of individuals. Among the prohibited artificial intelligence (AI) applications listed in article 15 of the European Commission's proposal for a Regulation on Artificial Intelligence \cite{EU-ai-act}, are applications that perform manipulative or exploitative online practices producing physical or psychological harms to individuals or \emph{exploit their vulnerability} on the basis of age or disability.
 
 The use of ML models that exploit or impact the vulnerability of persons can become an ethical and potentially legal issue, because it could lead to manipulation, reduced agency/autonomy, or altered behaviour of individuals, in a way that is not necessarily in their best interest. Currently it is difficult to know whether a given model's decisions are based on vulnerability-related information. However, public mandates for AI auditing must address this issue, and explainable AI (XAI) methods may be necessary for detecting vulnerability exploitation by ML based recommendation systems.

In this paper we are concerned with the problem of vulnerability detection by machine learning models, and our specific contributions are the following.
\begin{enumerate}
\item We define and describe examples of potential vulnerabilities and how they manifest in behaviour. 
\item We  argue, by means of a literature review, that machine learning models can and are being used to detect vulnerabilities.  
\item We illustrate cases in which vulnerabilities are likely to play a role in algorithmic decision making based on behavioural and contextual data. 
\item To ameliorate the current situation, we further propose a set of requirements that methods must fulfil in order to:
\begin{itemize}
\item detect the potential for vulnerability modelling, 
\item detect whether vulnerable groups are treated differently by the model, and 
\item detect whether an ML model has created an internal representation of vulnerability. 
\end{itemize}
\end{enumerate}
The paper structure follows the above order of contributions. We conclude with an outlook perspective on the importance to not only regulate applied AI systems on a technical level, but to put the human at the centre of this process \cite{Shneiderman2020}.

\section{Vulnerability in the context of machine learning}\label{sec:state}
In the context of algorithmic decisions systems, it is challenging to draw the line between persuasive  systems and manipulation. It is not simple to identify where free will ends and where manipulation starts  \cite{Mathur2021,Klenk2020} Hence, to understand the possible impact that machine learning systems can have on individuals from vulnerable groups, we need to establish a definition of vulnerabilities and create an understanding of how machine learning with big behavioural data can model these. 

\paragraph{Vulnerability}
The United Nations define vulnerabilities as ``The conditions determined by physical, social, economic and environmental factors or processes which increase the susceptibility of an individual, a community, assets or systems to the impacts of hazards.''\footnote{\url{https://www.undrr.org/terminology/vulnerability}}. Unlike in physical environments, where physical disabilities and vulnerabilities are impactful (e.g., paraplegia), in digital environments, psychological vulnerabilities (e.g., depression) are more exposed to exploitation and discrimination. Hence, here we are primarily concerned with psychological and social factors which increase the susceptibility of an individual to the impacts of hazards. 

When considering  psychological vulnerabilities, it can be helpful to distinguish between stable (i.e., traits; anxiety disorder) and momentary characteristics (i.e., states; momentary anxiety) of a person. In that regard, we define vulnerabilities for the purposes of this article as follows: 
\begin{description}
\item \noindent A person is vulnerable if their permanent or momentary psychological characteristics makes it particularly difficult for them to make autonomous choices and to exert agency over their actions under full consideration of their consequences and outcomes.
\end{description}

It is ``the quality or state of being exposed to the possibility of being attacked or harmed, either physically or emotionally.'' \cite{lexico}. For this reason, vulnerable individuals can be exposed to additional forms of harm or an aggravation of their condition (e.g., aggravation from mild depression to major depression) without their awareness and the agency to prevent or counteract it.

For the purpose of this article, we will use symptoms of major depression \cite{Dsm5} as an illustrative example of a relatively stable psychological vulnerability for which first evidence suggests the possibility to model it using machine learning and digital footprints \cite{Liu2022,Mueller2021}. The identification of people who are suffering the symptoms of major depression can in certain contexts be in violation of several ethical principles.

\subsection{Exploitable vulnerabilities in algorithmic decision systems}
In machine learning systems, vulnerabilities can be automatically exploited when the vulnerable state or condition of an individual becomes entangled with the optimisation criteria of an algorithmic system (i.e., recommender system). For example, depressive users on a social media platform might engage more with content that is emotionally charged with feelings of sadness, depression, and hopelessness. This behavioural tendency or ``interest'' might be picked up by the optimisation algorithm of that platform that is designed to maximise user engagement based on behaviour. Consequently, the platform could provide an increasing number of similar content to the user. While this optimisation procedure is effective to optimise user-engagement and harmless for non-vulnerable users (e.g., providing increasingly specific content on taco recipes), the increased provision of depression-related content to users with the vulnerable condition might aggravate the depressive symptoms of the user (e.g., reinforcement of perception of worthlessness) while increasing their engagement with the platform \cite{TikTokWSJ}. 

It is difficult -- and outside our aim -- to analyse the underlying causal patterns between mental states and compulsive behaviours. There is, however, sufficient reason to consider that emotions of depression, anxiety,  individuals having a negative view of themselves, coincide with mental conditions that lead to compulsive behaviours. 
For example, \cite{lejoyeux1997study} conclude that ``Compulsive buying is frequent among bipolar patients going through a manic phase. In most cases, the behaviour is associated with other impulse control disorders or dependence disorders and a high level of impulsivity''. Similarly, \cite{lejoyeux2002impulse} conclude that ``Our data emphasises the frequency of association between ICDs (impulse control disorders) and major depression, and 29\% of the depressed patients also had an ICD''. Shopping as a coping behaviour for stress also is investigated in \cite{yasuhisa2001}, who report that ``more stress release was found with larger amounts spent''.
\cite{mcelroy1996impulse} state that ``Although no studies directly compare a cohort of ICD patients with a cohort of mood disorder patients, available data suggest that ICDs and bipolar disorder share a number of features: (1) phenomenological similarities, including harmful, dangerous, or pleasurable behaviours, impulsivity, and similar affective symptoms and dysregulation \dots''. One could argue that these are exactly the kind of behaviours social media usage is likely to reflect, probably even before they are discovered clinically or by the individuals themselves~\cite{facebook_depression2018}.

This insight must be combined with the knowledge that data based models for identifying and predicting mental health conditions are used for commercial purposes. For example, a leaked Facebook document reported by The Australian \cite{arstechnica_teens2017} revealed that the platform uses data based models to identify young, meaning down to $14$ year old, individuals feeling vulnerable, i.e. ``worthless'', ``insecure'', ``stressed'', ``defeated'', ``overwhelmed'', ``anxious'', ``nervous'', ``stupid'', ``silly'', ``useless'', and ``a failure''. Furthermore, the document, marked ``Confidential: Internal Only'', outlines how Facebook can target ``moments when young people need a confidence boost'', and reveals an interest in helping advertisers target moments in which young users desire ``looking good and body confidence'' or ``working out and losing weight''.

As stated by \cite{inkster2016decade}: ``A key ethics challenge for using social networking site data (\dots) will be to ensure that vulnerable individuals have a comprehensive and sustained understanding of what participation involves\dots''. We fully agree with this conclusion, and argue that most users of social platforms do indeed \textit{not} have such a comprehensive and sustained understanding. Sadly, this is neither a novelty nor a controversial stance. However, we also argue that this problem exists on two levels, one level being the mere flow of information, or desire of social media platforms to inform users of how their data is being used. The second level is more subtle as well as technical, since ``understanding of what participation involves'' requires an understanding of the models used to analyse the data of the individual. Such understanding of non-interpretable models is at present often not possible, commonly referred to as the ``black box'' problem in machine learning and artificial intelligence.

This must be considered in light of the observation that the reviews quoted in the beginning of this section -- \cite{wongkoblap2017researching,aboureihanimohammadi2020identification}, reporting the general tendency of machine learning methods replacing traditional forms of data analysis -- consequently report an increase in \textit{non-interpretable} models being proposed to predict psychological constructs and to detect mental health disorders.

All these things can be considered vulnerabilities, and in section~\ref{sec:digbehdat} we summarise research suggesting that AI can be used to detect such vulnerabilities.
\subsection{Digital behavioural data for assessing psychological constructs}\label{sec:digbehdat}
Unlike physical characteristics (e.g., body height), latent psychological characteristics of individuals cannot be measured directly but need to be estimated from reports, observations, or psychometric tests (e.g., cognitive abilities) of peoples thoughts, feelings, and behaviours. For example, most forms of mental health disorders are diagnosed via standardised self-report scales and structured clinical interviews. However, the self-reported information that is collected in this diagnostic process is highly subjective and retrospective \cite{martin1990mental}. This is problematic because many psycho-pathological conditions impact peoples cognitive abilities (e.g., concentration, memory), hence potentially bias the obtained information. Moreover, self-reports are subject to a myriad of methodological influences \cite{Vaerenbergh2013} and intentional faking \cite{Goerigk2020}.

New approaches have been developed to assess psychological phenomena and psychopathology combining objective data on people`s digital behavioural and online footprints with machine learning \cite{Insel2017,Stachl2021PS}. Text data in particular has long been considered as a valid source of information on psychological processes \cite{Allport1942}. As early as 1982  \cite{oxman1982language} reported that patients could be classified into groups suffering from depression and paranoia based on linguistic speech analysis. More recent work \cite{Azucar2018,Kosinski2013} has demonstrated that personal characteristics and traits can be inferred from text and other digital footprints on social media platforms. Others have demonstrated that these data can be used to predict individual well-being \cite{Schwartz2016PredictingIW} and future mental illness \cite{thorstad2019predicting}. 

Particularly well explored is depression \cite{de2013predicting,schwartz2014towards,orabi2018deep,facebook_depression2018,tadesse2019detection,arabic_depression2020}, including comorbidity such as self-harm \cite{Yates2017DepressionAS}, and anorexia \cite{ramiirez2018upf}. Quantifiable signals in Twitter data relevant to bipolar disorder, major depressive disorder, post-traumatic-stress disorder and seasonal affective disorder were demonstrated by \cite{coppersmith2014quantifying}, who constructed data-based models capable of separating diagnosed from control users for each disorder. These systematic reviews, \cite{wongkoblap2017researching,aboureihanimohammadi2020identification,10.1145/3398069}, provide an overview of these approaches.

Another growing body of research suggests that more fine-grained behavioural and contextual data that can be collected with of-the-shelf smartphones allow for similarly accurate predictions of psychological phenomena with much smaller samples \cite{Stachl2020PNAS,Panicheva2022}. Computational inferences from mobile sensing data include a number of psychologically-relevant individual differences including demographics \cite{Koch2022,Malmi2016,Sundsoy2016}, moral values \cite{Kalimeri2019}, and personality traits \cite{Stachl2020PNAS}.
Latest work, has started to explore the feasibility to predict clinical depression levels using messaging texts and sensor readings from smartphones \cite{Liu2022,Mueller2021}.
Finally, efforts to recognise unstable psychological states such as affective-emotional states \cite{Israel2020} or cognition \cite{Gordon2019,Rauber2019} proved more challenging and constitute an area of ongoing research. The algorithmic recognition of these states is highly relevant to the understanding of more complex psychological phenomena such as depression \cite{Liu2022}, yet more difficult to achieve.

Extant evidence sufficiently suggests that information about psychological traits and mental health states can be linked to digital behavioural footprints. However, it should be noted that despite a large body of research literature on the subject, obtaining \textit{clinically valid} diagnostic information on individuals is challenging. This, as discussed in \cite{mental_health_some2019}, leads to subsidiary ``proxy diagnostic signals'', meaning characteristic online behaviours, being used instead in this kind of research. \cite{mental_health_some2019} also report that these diagnostic signals lead to models with poor external validity, cautioning against their use for clinical decision-making.

While data based tools assessing mental health can certainly be used to help vulnerable individuals -- suicide prevention tools have for instance been available on Facebook for more than ten years \cite{facebook_suicide_prevention} -- their existence also raises several concerns: The general desirability of subsequent interventions has not been democratically agreed upon. Whether the use of data for this purpose can really be characterised as voluntary for Facebook users is not clear, as evidenced by e.g.\ the public dispute of the study `Experimental Evidence of Massive-Scale Emotional Contagion Through Social Networks' \cite{Kramer8788}, including its editorial expression of concern. Showing that social media data can be used for emotional contagion -- \textit{``leading people to experience (\dots) emotions without their awareness''} --, this study together with the existing body of literature lead to the conclusion that not only \textit{can} data and machine learning models be used to detect emotional states and mental health conditions, but that such models are already being successfully developed and used to change them.

Due to this fact caution is warranted in the way how these normally protected information is treated in digital environments.

\section{Towards detecting vulnerability exploitation in machine learning models}\label{sec:approach}
As evidenced in the previous section, ML models are being successfully developed with the purpose of modelling vulnerability. Thus, we must consider whether models developed for other purposes -- for example targeted advertisement -- on data that contains information about vulnerability, are in fact modelling and thus exploiting this information incidentally.

As discussed in~\cref{sec:state}, there is no single marker for vulnerability: it is not directly represented by data features, and must be deduced. While age is something that can be a data feature, human beings do not have e.g.\ \vulnerability{} written somewhere on their body, and no single behaviour is uniquely linked to \vulnerability{}. Furthermore, vulnerability is context dependent. Therefore, in order to detect vulnerabilities, one would have to test for specific ones, and given the relevant context. To illustrate, we provide a list of examples of vulnerabilities in their contexts, see~\cref{tab:vulnerabilities}. This list is certainly not exhaustive, and merely included for illustration. A complete list of vulnerabilities to test for would have to be created by legislative operators in the respective administrative region (e.g., the European Union) and kept updated by national agencies and government organs charged with data and/or consumer protection.

\begin{table}[H]
    \caption{Non-exhaustive list of potential vulnerabilities and corresponding relevant contexts. Abbreviations: Post traumatic stress disorder (PTSD); Generalized anxiety disorder (GDA).}
    \label{tab:vulnerabilities}
    \resizebox{\textwidth}{!}{
    \centering
    \begin{tabular}{ll}
        \toprule
        \textbf{State} & \textbf{Context} \\
        \midrule
        Low impulse control (e.g., in Schizophrenia) & In-game purchases \\
        Excessive feelings of anger and violence (e.g., PTSD) & Continuous violence media, trolling \\
        Extreme mistrust (e.g., in Schizophrenia) & Conspiracy theory consumption \\
        Excessive fear or worries (e.g., in GAD) & Continuous negative news in feed (doomsday scrolling) \\
        Feeling sad or down & Shopping as compensation \\
        Low cognitive abilities/intelligence & Susceptibility to logically flawed arguments (disinformation) \\
        Increased risk taking behaviour (e.g., in bipolar disorder) & Online gambling \\
        Narcissistic personality disorder & Environment giving recognition for appearance \\ & or activities (e.g.\ social media) \\
        \bottomrule
        \end{tabular}}
\end{table}

Psychological conditions that can form the base for vulnerabilities are ascertained by psychologists who in turn have to use self-reported feelings and behavioural tendencies to find out whether a person has a mental illness or vulnerability. Alternatively, machine learning models can use data containing language or behavioural patterns to construct latent features representing mental illness or vulnerability in a person. However, if a model is not specifically constructed to detect a vulnerability, how can we be sure that it hasn't constructed latent features representing vulnerabilities, and exploits these to achieve its goal?

In order to understand whether and how a machine learning model incorporates vulnerabilities, we need to be able to answer some questions about this suspected model. It should also be clear who is able to understand the relationship between the predictions of the model and a specified vulnerability. In addition, there needs to be an understanding of how to sufficiently empower concerned stakeholders to enforce necessary changes to the suspected model and to mitigate the creation and the use of ethically undesirable models in general. Namely, not to simply impose restrictive regulations that prohibit certain model use, but rather to create tools to detect and measure what models do wrong and when. 
\subsection{Information contained in data}
\begin{question}
    Is sufficient information about a given vulnerability contained in the training data of the model?
\end{question}

A data based model consisting only of operations on the input data -- which is the case for all ML models, including neural networks -- cannot add information beyond what is present in the input data. This is formalised in the Data Processing Inequality \cite{Beaudry2012AnIP}, stating that post-processing cannot increase information. Therefore, the necessary information for detecting a vulnerability must be present in the data available to the model under scrutiny. If it is, then there is a possibility that the model uses this information, \emph{although this might not be the prime objective of the model}.

\begin{case}
    In order to model the manic phase in a bipolar disorder, the data used by the model must contain either direct or indirect information about manic behaviour. 
\end{case}

Information about mental states and vulnerability can either be directly available (e.g., questionnaire data, estimates from models), or indirectly through correlation structures in the data. For example, depression symptoms can be characterised by feelings of sadness, fearfulness, emptiness or hopelessness -- information that is frequently expressed in self-statements (e.g., text messages, see \cite{Liu2022}). Depressive episodes can also be indicated by more basic behavioural patterns such as unusual diurnal activity (i.e., sleep irregularities) or reduced physical activity, which can also be reflected in digital behavioural data \cite{Schoedel2020,Saeb2016}.
Knowing common symptoms of depression (e.g.\ from \cite{Dsm5}), we can flag data sets containing information about these behaviours as disposed for modelling depression. 

\begin{figure}
    \centering
    \includegraphics[width=0.7\columnwidth]{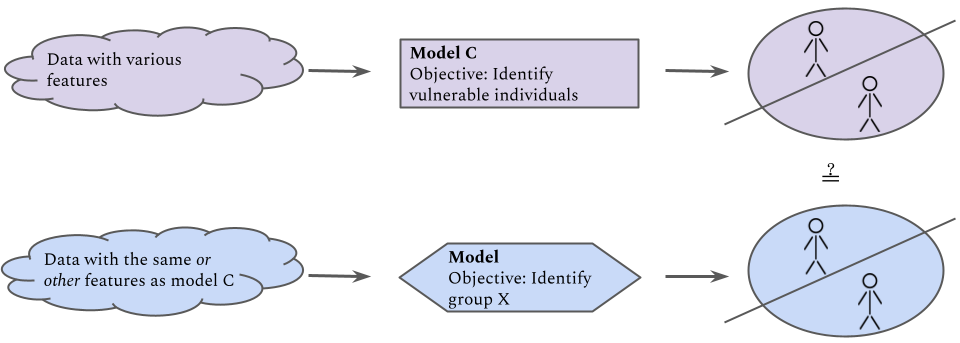}
    \caption{\label{fig:two_models} The model $C$ (top, purple), known to identify vulnerable individuals, can be used to test whether the model under scrutiny (bottom, blue) is likely to have modelled vulnerability in individuals in order to achieve its objective.}
\end{figure}

\subsection{Model behaviour}
\begin{question}
    Are vulnerable groups systematically treated differently by the model?
\end{question}
Assuming we know the true vulnerability status of individuals, we can either monitor the predictions of the model under scrutiny, or attempt to directly compare the model to one that is known to incorporate the vulnerability of interest. The former is done by systematically analysing the predictions of the model and investigate whether there is a high correlation with the predictions of vulnerability. 
\begin{case}
Assume that we have a data set containing true labels of low impulse control in individuals, and want to investigate whether a targeted advertisement recommender model exploits the vulnerability of these individuals to cause compulsive buying behaviour. Then, we could let the recommender model predict on the data describing the low impulse control individuals as well as a control group consisting of individuals not sharing the vulnerability, and compare the scores assigned to all individuals by the model. If a significantly higher number of individuals with low impulse control are identified as potential buyers by the recommender model, then this model is likely to exploit knowledge of the low impulse control in its prediction. This approach is possible in the case of compulsive buying, since we know that it is possible to predict compulsive buying from pathological personality traits, see e.g.~\cite{Harnish2021}.
\end{case}

The second approach relies on the availability of a model that intentionally predicts the vulnerability of interest and that could serve as a ``ground truth'' source. Here, we give an example to illustrate this approach.
\begin{case} 
We can train a model $C$ to detect \vulnerability{} in individuals, and construct a test data set consisting of both individuals having \vulnerability{} as well as non-afflicted individuals. We can then compare directly how many of the individuals assigned the vulnerability label by $C$ are also identified by the model under scrutiny, either by being assigned to the same class (in the case of classification) or given similar scores (in the case of regression). This method is visualised in~\cref{fig:two_models}. Note that the features each model is trained on need not be the same in order to give the two models the same information about an individual's vulnerability; most behavioural data are proxy variables.
\end{case}

\subsection{XAI and existing methods}\label{sec:xai}
\begin{question}
    Has the model built an internal representation of vulnerability?
\end{question}

This approach is introspective, i.e.\ it involves investigating the internal parameters of the model in order to infer whether these contain sufficient information to separate vulnerable from non-vulnerable individuals. 
Restricting ourselves to neural networks, we can use methods allowing us to probe the internal state of a neural network for concepts, as described in \cite{kim_tcav}. 
Interpreting neural networks' latent features as human concepts has been studied for a variety of data and model types, including for words \cite{NIPS2013_words}, images \cite{concepts_gans}, and chess \cite{alphazero_concepts}.

\begin{case}
If we regard a specific vulnerability, for instance \vulnerability{}, as a concept, we can use concept activation methods to find out whether a neural network model has internalised a representation of this vulnerability as follows.
A labelled data set consisting of people with and without the \vulnerability{} affliction is assembled, and used as positive and negative probes for the concept. Providing this data to the neural network, we collect the activations of its internal layers, and investigate whether these activations can be used to separate the \vulnerability{} afflicted individuals from the non afflicted individuals in the test data set. 
The accuracy with which this can be done indicates to which degree the neural network has represented the vulnerability concept. If we can identify representation of a concept by a late layer inside the neural network, or throughout the entire model, we can assume that it has found this information useful and  makes use of it for its  its classification.
\end{case}

\begin{question}
    Does the model assign high importance to features associated with a vulnerability?
\end{question}

Feature importance attribution methods can be used in order to determine which features have a high impact on a model's prediction, i.e.\ which features are ``perceived as important by the model itself''. Popular libraries for generating feature attributions include SHAP \cite{Lundberg2017AUA}, SAGE \cite{covert_sage}, and LIME \cite{lime}. Based on the resulting feature importances, a domain expert on human psychology can assess whether the model makes use of features that are informative with respect to a vulnerability. 

\subsection{Fairness and vulnerability}
The concern that a decision-making algorithm, regardless of whether it is model-based or rule-based, treats disadvantaged groups of people differently, falls within the domain of fairness in AI \cite{barocas-hardt-narayanan}. 
Fairness in AI is concerned with protecting the ethical value of justice. The justice principle is concerned with how people are treated and embodies the idea that decisions about individuals should be based on just arguments, the consequence of which is that similar people should be treated similarly. Group fairness in particular is concerned with ensuring that decisions to recommend an allocation of a resource, such as a job interview, are not directly or indirectly biased with respect to a legally protected feature of the applicant, for example race. 

The fundamental question of fairness is \emph{``Is a particular group in treated unjustly or harmed?''}. The group in question can be defined by protected attributes, such as race, religion, gender, etc. In principle, groups can also be characterised by vulnerability, and group fairness methods can be applied to issues of detecting vulnerability and de-biasing of a model used in a decision-making algorithm. In practice this might be difficult.

In order to use fairness metrics to detect vulnerability bias in decision-making, knowing the exact set of features that characterise that vulnerability is necessary. The set of features that characterise a protected group is always given, typically defined by law. On the other hand,  for many mental health conditions, a clear set of diagnostic features is not available.  

Most group fairness metrics assume the existence of a desirable model classification and measure the proportionality of two groups being classified. The de-biasing methods for ML models are a collection of methods that deploy different changes in the decision-making pipeline, which result in enforcing proportionality in desirable and undesirable decisions across two groups \cite{Belamy2018}. At first glance, such methods could be shoehorned to change the proportion of vulnerable group members vs others that are, for example, targeted with a particular link. But these methods are effective in adjusting small differences and trading off model accuracy for group proportionality. If the overlap between vulnerable people and targeted people is large, current de-biasing methods would not be effective. 

\section{Challenges and call to action} \label{sec:shallenge}
We argue that methods for determining \emph{whether} a model is in fact exploiting vulnerabilities must be developed and integrated in auditory frameworks. In this section, we first address challenges associated with developing such methods in \cref{sec:challenge_data,sec:challenge_model}, before pointing to the protections for vulnerable groups in specific regulation in \cref{sec:challenge_microtargeting}.

\subsection{Data describing vulnerabilities} \label{sec:challenge_data}
In order to apply the model monitoring, model comparison or concept detection approaches described in Examples 2, 3 and 4 respectively, labelled data sets describing vulnerable individuals would have to be created. This immediately poses an ethical challenge: we would have to decide whether collecting and labelling data describing vulnerabilities is unethical, \emph{although} the purpose is to identify and regulate models that do this. As we regard the creation, storage and use of such data sets as too risky to defend the potential gain, we wish to point out a possible alternative venue of research involving synthetic data. It could be possible to generate synthetic data, based on psychological descriptions of vulnerability traits. Similar venues are currently being explored in the context of medical data, to facilitate training ML models for clinical decision support without having to collect and store medical data \cite{deepfake_ecg}, although not without challenges of its own \cite{chen2021synthetic}.

\subsection{Intentionally modelling vulnerability}\label{sec:challenge_model}
There are also challenges associated with the creation and use of models trained to identify a state that can constitute a vulnerability, i.e.\ with the approach described in Example 2 and creating ``model $C$''. The development of such models requires diagnostic, meaning sensitive and privacy protected, information about individuals. Hence there are many challenges and professional requirements that need to be met for safe and ethical handling and development of such models~\cite{pargent2022}.

It has been argued\footnote{See for example~\cite{medium_stereotypes} for the discussion regarding ML use in detecting sexual orientation.} that models with such capabilities should not be developed due to the risk of abuse. However, we argue that the existence of models accidentally or covertly having the capabilities of identifying vulnerabilities constitutes a far greater risk of abuse -- understanding something is often better than dogmatically banning it. This opens up a discussion on proportionality which is common in privacy law, and outside the scope of this paper.
Still, we wish to point out the interesting trade-off that arises as developing type $C$ models helps detect accidental vulnerability modelling, while focusing the attention on difficult questions such as ``Who should develop type $C$ models? Who should be trusted with using type $C$ models to test models already deployed? Can type $C$ models be distributed, or does their containing latent features describing vulnerability force us to consider them as containing sensitive information?

We most adamantly do not intend to argue that all platforms using models trained on publicly available behavioural data should also ask their users to provide sensitive data about their mental health status, in order to develop type $C$ models for testing. 
\subsection{Microtargeting and regulation}\label{sec:challenge_microtargeting}
The  approaches we describe in section~\ref{sec:approach} require access to the models themselves in addition to labelled data containing vulnerable individuals.
Researchers and private initiatives typically do not have access to the commonly used recommender systems nor to the necessary data. 
Since many of the models that should be tested for vulnerability modelling are proprietary, we cannot require or expect that research groups be given access to these. It is only regulatory and supervisory authorities that can require such access in auditory processes. However, an effective regulatory framework for the protection of vulnerable individuals in AI may be hard to attain. The core problem with establishing and enforcing regulation is articulating the concern: what are the issues that we would like to avoid and where do they occur.

One area in which there is an articulated concern, and some legal protection, for the identification and  exploitation of vulnerable states is targeted advertising, or rather its specific digital version called `microtargeting'.

Targeted advertising is advertising directed at an audience with certain characteristics. Although this type of marketing has always existed, the availability of data traces and online advertising has created the option to {\em microtarget} audiences and customers. Microtargeting is the practice of  using data mining and AI techniques to infer personal traits of people and use them to adjust the marketing campaign to those traits. Micro-targeting  gives advertisers the ability to exploit recipients' personal characteristics and potential vulnerabilities \cite{Lorenz-Spreen2021}. 

It is not entirely clear how effective micro-targeting is in persuading individuals \cite{Rafieian2021,tappin2022}. What is clear is that micro-targeting at present is not transparent as to what data it uses to tailor exposure to content. If exploitation of vulnerabilities is happening, there are no mechanisms to detect it and mitigate it \cite{fare2022,Lorenz-Spreen2021}. 

Online targeted advertisement is subject to data related and regulations in the European Union. \cite{Covington2022} identify  the following rules on online targeted advertising:
\begin{itemize}
    \item the ePrivacy Directive (Directive 2002/58/ED, as amended);
    \item the GDPR (Regulation (EU) 2016/679, as amended);
    the eCommerce Directive (Directive 2000/31/EC);
    \item the Unfair Commercial Practices Directive (Directive 2005/29/EC, as amended);
    \item the Directive on Misleading and Comparative Advertising (Directive 2006/114/EC, as amended);
    \item the Audiovisual Media Services Directive (Directive (EU) 2018/1808); 
    \item the Consumer Rights Directive (Directive 2011/83/EU, as amended);
    \item The Digital Markets Act (``DMA''); and
    \item The Digital Services Act (``DSA'') 
\end{itemize}

Most of these regulations are concerned with obtaining the informed consent from a user for the processing of their data, as well as clearly indicating which legal or natural person has commissioned the advertising. The Audiovisual Media Services Directive prohibits the use of surreptitious or subliminal techniques of advertising when those cannot be readily recognised as such, however, micro-targeting is not considered a surreptitious or subliminal technique. 

Most regulations are not explicitly concerned with vulnerable individuals. An exception is the recently agreed Digital Services Act (DSA). Article 63 specifies: 
\begin{quote}
The obligations on assessment and mitigation of risks should trigger, on a case-by-case basis, the need for providers of very large online platforms and of very large online search engines to assess and, where necessary, adjust the design of their recommender systems, for example by taking measures to prevent or minimise biases that lead to the discrimination of persons in vulnerable situations, in particular where this is in conformity with data protection law and when the information is personalised on the basis of special categories of personal data, within the meaning of Article 9 of the Regulation (EU) 2016/679. In addition, and complementing the transparency obligations applicable to online platforms as regards their recommender systems, providers of very large online platforms and of very large online search engines should consistently ensure that  recipients of their service enjoy alternative options which are not based on profiling, within the meaning of Regulation (EU) 2016/679, for the main parameters of their recommender systems. Such choices should be directly accessible from the interface where the recommendations are presented.
\end{quote}

Concern and request for protection for vulnerable recipients occurs twice in the DSA, in both case minors are used as an example of such group of recipients. 
\section{Conclusion}
In the very same way that specific vulnerabilities can lead to a feeling of being overwhelmed in the affected person, it seems that machine learning engineers, policy makers, and ethics specialists shy away from addressing them in the design and audit of autonomous learning systems. Both conditions need to be ameliorated. We are only now beginning to explore the power of behavioural prediction models and their potential for monetising. 

To to protect vulnerable individuals and groups, three essential conditions must be met: i) understanding how vulnerable states can be exploited, ii) detecting when information about vulnerability is being used by an algorithm, iii) and creating social and financial incentives for preventing exploitation of vulnerabilities. 


The study of the relationships between individuals' vulnerabilities and their online behaviours, which may be exploited or monetised, falls within the realm of the behavioural sciences. This research can lead to the development of technology and laws that help prevent exploitation, but it also has the potential to aid exploiters in their efforts. One often proposed solution to this dilemma is to prohibit research that aims to understand such correlational patterns. However, this approach also has drawbacks, as it may prevent the advancement of knowledge and the creation of interventions to help vulnerable individuals.

Around the clock observation can be a privilege in monitoring a condition for those who can afford it (e.g., relapse prediction). It can also be a tool for oppression through surveillance. Combined with modern tools offering cheap and consistent unobtrusive observation, ML can help us understand mental health conditions that we otherwise would not be able to afford. Dishonest actors will always be able to identify exploitable vulnerabilities because complete and outright prohibition of methods whose impact we do not fully understand is not feasible. 

We propose several approaches to identifying if an ML model uses information related to vulnerability. It is difficult to evaluate the approaches we propose because data sets to do so are not available. In domains in which we can easily suspect information on vulnerabilities is used, such as targeted advertising, the information is on who is being offered which content is understandably hard to come by \cite{fare2022}. One approach would be to consider proxy domains, such as recommendations of movies or songs and look into who is being recommended what type of entertainment. This is what the \cite{TikTokWSJ} investigation attempted.  We would still be left with the problem of having access to the information of the actual mental state of individuals.  This is not information that should be publicly available. This is why iii) is not only needed but necessary. 

Advertising is an example of a context in which we can clearly see how vulnerabilities can be monetised and exploited. It is not necessarily the only such example. And even in this context, regulation is hard to agree on and enforce in time. In order to make AI safe and trustworthy, it is imperative that regulators engage in a debate and collaborate with experts in AI ethics, XAI and the behavioural sciences, to understand vulnerabilities and to regulate the domain.  
\paragraph{Acknowledgement. The contribution of C.S. was partially funded by a research collaboration with armasuisse (contract: 8203004934) and by an internal University of Bergen grant for visiting researchers.}
\bibliographystyle{theapa}
\bibliography{bibliography}
\end{document}